\documentclass{article}

\usepackage{arxiv}

\usepackage[utf8]{inputenc} 
\usepackage[T1]{fontenc}    
\usepackage{hyperref}       
\usepackage{url}            
\usepackage{booktabs}       
\usepackage{amsfonts}       
\usepackage{nicefrac}       
\usepackage{microtype}      
\usepackage{lipsum}
\usepackage{graphicx}

\usepackage{bm}
\usepackage{amsmath,amssymb,amsfonts}
\usepackage{subcaption}
\usepackage{enumitem}
\usepackage{multirow,multicol}
\usepackage{arydshln}

\newcommand{\argmin}{\mathop{\mathrm{arg~min}}\limits}
\newcommand{\C}[1]{$C_{\mathrm{#1}}$}

\title{Fast instance-specific algorithm configuration with graph neural network}

\author{
Shingo Aihara \\
    AI Laboratory\\
    Fujitsu Limited\\
    Kawasaki 211-8588, Japan\\
    \texttt{aihara-shingo@fujitsu.com} \\
\And
Matthieu Parizy \\
    AI Laboratory\\
    Fujitsu Limited\\
    Kawasaki 211-8588, Japan\\
}

\begin{document}
\maketitle

\begin{abstract}
    Combinatorial optimization (CO) problems are pivotal across various industrial applications, where the speed of solving these problems is crucial.
    To enhance the performance of solvers for CO problems on diverse input instances, it is essential to fine-tune the solver parameters for each input of those instances.
    However, this tuning process is time-consuming, and the time required increases with the number of instances.
    To address this, a method called instance-specific algorithm configuration (ISAC) has been devised.
    This approach involves two main steps: training and execution.
    During the training step, features are extracted from various instances and then grouped into clusters.
    For each cluster, parameters are fine-tuned.
    This cluster-specific tuning process results in a set of generalized parameters for instances belonging to each class.
    In the execution step, features from an unknown instance are extracted to determine the class.
    Once the cluster is identified, the corresponding pre-tuned parameters are applied.
    Generally, the running time of a solver is evaluated by the time to solution ($TTS$).
    However, methods like ISAC require preprocessing, which adds to the total execution time ($T_{tot}$).
    Therefore, $T_{tot}$ is the sum of $TTS$ and the tuning time ($T_{tune}$), expressed as $T_{tot}\ (=TTS+T_{tune})$.
    While the goal is to minimize $T_{tot}$, it is important to note that extracting features in the ISAC method requires a certain amount of computational time.
    This research proposes a method to streamline the computationally intensive feature extraction and class determination processes in the execution step of the ISAC method using a more efficient classifier to significantly reduce computational time.
    The features used in the experiment include summary statistics of the solution update history and the set of satisfying solutions obtained during the solver execution, which takes several 10 seconds.
    The classifier is constructed using a graph neural network.
    Experimental results show that $T_{tune}$ in the execution step, which take several 10 seconds in the original ISAC manner, could be reduced to sub-seconds.
\end{abstract}

\bigskip


\section{Introduction}

Combinatorial optimization (CO) problems are integral to numerous fields, including transportation, logistics, finance, and manufacturing.
These problems can be formulated as different types, such as the knapsack problem and the traveling salesman problem (TSP)~\cite{Springer2012CombinatorialOptimization}, and solving them quickly is crucial in industrial contexts.
Many of these problems can be expressed as binary quadratic programming (BQP) problems, where the goal is to maximize or minimize an objective function subject to certain constraints~\cite{Springer2022QUBOProblem,AnnalsOR2022V314p141-183,fphy2014V2a5p1-15,DiscApplMath2007V155p623-648}.
Specifically, BQP with inequality constraints, such as the knapsack problem, is defined as follows:
\begin{equation}
    \label{eq:bqp_problem}
    \begin{split}
        \text{minimize}\quad & \bm{x}^T \bm{Q} \bm{x} \\
        \text{subject to}\quad & \bm{A}\bm{x} \leq \bm{b} \\
        & \bm{x} \in \{ 0, 1 \}^n,
    \end{split}
\end{equation}
where $\bm{x} \in \mathbb{B}^n$ is an $n$-dimensional binary variable vector, $\bm{Q} \in \mathbb{R}^{n\times n}$ is a real-valued $n{\times} n$ symmetric matrix, $\bm{A} \in \mathbb{R}^{m\times n}$ is a real-valued $m{\times} n$ matrix, and $\bm{b} \in \mathbb{R}^m$ is a real-valued $m$-dimensional vector.

The BQP problem, where the number of candidate solutions grows exponentially with size $n$, makes it challenging to obtain the optimal solution in a realistic time frame.
To address this, recent research has focused on developing heuristic solvers, both in hardware and software, specifically designed to efficiently approximate solutions to BQP problems.
Notable examples include D-Wave's quantum annealing leveraging superconducting states~\cite{D-Wave_homepage}, Toshiba's SQBM+~\cite{Toshiba_SQBM+}, NEC's vector annealing~\cite{NEC_VectorAnnealing}, and Fujitsu's Digital Annealer (DA)~\cite{Fujitsu_DigitalAnnealer}.
These heuristic solvers operate using GPUs and dedicated accelerators and employ algorithms such as simulated annealing, genetic algorithms, and tabu search~\cite{Springer2022DesignOfHeuristicAlgorithms}.

Generally, the performance of heuristic solvers such as DA and classical mathematical optimization solver like SCIP~\cite{MPC2009V1p1-41} and GUROBI~\cite{gurobi} is evaluated based on the quality of the solution and the time taken to reach these solutions.
This performance significantly depends on the parameters of various functions implemented in their algorithms.
Therefore, it is crucial to fine-tune the parameter configuration to the specific characteristics of each problem~\cite{Springer2018HandbookOfMetaheuristics}.
Given the vast range of possible configurations, manual exploration or grid search is impossible, prompting the development of algorithm configuration (AC) as a method for automatic and efficient parameter exploration.
The AC method obtains an optimal parameter $\theta^*$, within the parameter space $\Theta$, for a certain objective function $F_{\mathcal{T}}$ given a specific instance, and is expressed as follow:
\begin{equation}
    \theta^* = \argmin_{\theta\in\Theta} F_{\mathcal{T}}(\theta).
\end{equation}

AC methods can be broadly classified into model-free and model-based approaches, depending on whether they use a surrogate model to predict search performance.
Model-free methods include examples like the gender-based genetic algorithm (GGA)~\cite{CP09p142-157}, which employs two distinct crossover strategies to identify parameters that demonstrate strong performance, and irace~\cite{ORP2016V3p43-58}, which updates the probability distribution of high-performance parameters.
Model-based AC methods often employ the sequential model-based global optimization (SMBO) algorithm.
SMBO constructs a surrogate model of the solver performance by analyzing past performance data, then proposes the next observation point to minimize the surrogate function derived from this model.
Prominent methods include those that use Gaussian processes (GPs) as surrogate models, such as Optuna~\cite{ProcKDD2019p2623-2631} and hyperopt~\cite{ProcICML2013V28p115-123}, which employ the tree-structured Parzen estimator (TPE)~\cite{NIPS2011V24p2546-2554}, and the sequential model-based algorithm configuration~\cite{LION2011p507-523} method that relies on Random Forest (RF).

When tackling CO problems, such as the TSP, users often face multiple instances of similar specific problems.
Tuning solver parameters for each instance individually is inefficient.
Therefore, a method has been devised to enhance performance by executing an AC method using multiple instances within a specific distribution~\cite{LION2011p507-523,ICCE2023p1-6,ACC24p98-103}.
This method allows for acquiring a generalized parameter configuration applicable to homogeneous instances within a specific distribution, significantly reducing tuning time.

In cases where instance distribution is heterogeneous, the standard AC method may not sufficiently improve solver performance.
To address this, the ISAC approach~\cite{ECAI2010p751-756} was proposed, as illustrated an overview in Fig.~\ref{fig:Fig_ISAC}.
ISAC consists of two steps: a training step and an execution step.
During the training step of Fig.~\ref{fig:Fig_ISAC}-(A), instances are grouped into clusters based on the similarity of their features.
A model-free AC method, such as GGA, is then applied to each cluster.
Here, the instance features play an important role, with the original approach utilizing manually designed variables and contract numbers to determine these features.
In the execution step of Fig.~\ref{fig:Fig_ISAC}-(B), the features of a new instance are computed, and its class is determined based on its distances to cluster centers in the feature space.
The parameter configuration of the nearest cluster is then used.
A notable approach is the deep graph clustering-based algorithm configuration~\cite{EAAI2023V125p106740}, which employs a graph autoencoder to derive embedding vectors for clustering purposes.

With the rise of cloud computing, the scale and variety of problems have significantly expanded, posing challenges for experts to individually fine-tune solver parameters for each user's specific instances.
To address this, AC method for individual instances~\cite{ICCE2023p1-6} and for instances within a specific distribution~\cite{ACC24p98-103} have already been developed.
However, these tuning methods are not immediately applicable to new, unseen instances.
Therefore, the ISAC approach proves highly advantageous, pre-training on instances from diverse problem classes and identifying the appropriate class for a new instance at runtime.

This study aims to minimize the sum of the time-to-solution ($TTS$) and parameter tuning time ($T_{tune}$), together represent the time for pre-processing, including feature extraction, and solver execution, by utilizing the ISAC method.
When using the ISAC method, features are required in the execution step.
Features strongly related to solver performance are considered crucial, with actual solver operation logs readily accessible.
However, since the solver execution time for feature extraction requires a certain amount of time, it is counted towards $T_{tune}$ in execution step, resulting in a loss in overall time.
To address this, we propose extending the ISAC framework to manage the computation of time-intensive features, such as obtaining solver logs, by framing it as a classification problem.
This is achieved using a graph neural network (GNN), which can be computed in a short time during the execution step.

In summary, this paper makes the following contribution.
\begin{itemize}
    \item We propose to train GNNs with the solver performances and classify instances using the trained GNN model.
    \item We demonstrate significant time savings by replacing parameter selection with feature extraction, which consumes a lot of time in the execution step, with GNN classification.
\end{itemize}

This paper is organized as follows.
Section~\ref{sec:methodology} describes our method.
Section~\ref{sec:experimental_setup} shows conditions for the method and a data set for evaluation.
Section~\ref{sec:result_and_discussion} reports experimental results and discusses the results.
Section~\ref{sec:conclusion} summarizes this paper and discusses our future works.

\section{Methodology}
\label{sec:methodology}
\subsection{Overview}

\begin{figure}[htbp]
    \centering
    \begin{minipage}[t]{0.39\columnwidth}
        \centering
        \includegraphics[width=\columnwidth]{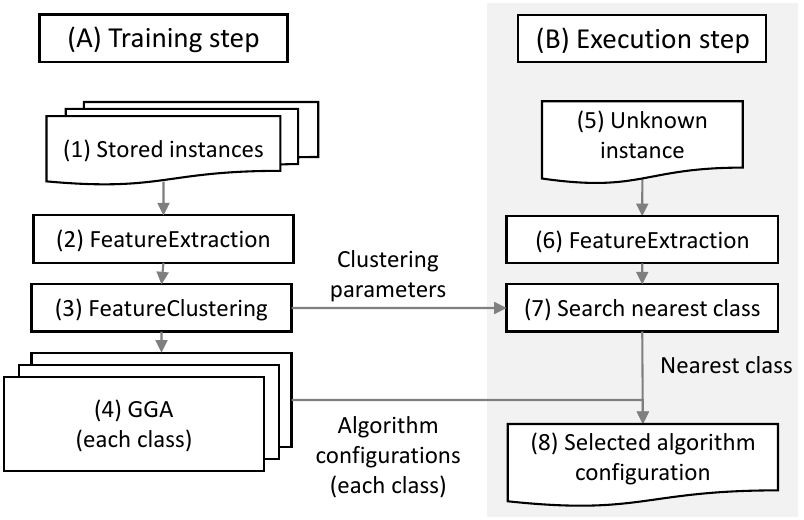}
        \caption{Overview of the instance-specific algorithm configuration framework.}
        \label{fig:Fig_ISAC}
    \end{minipage}
    \begin{minipage}[t]{0.59\columnwidth}
        \centering
        \includegraphics[width=0.95\columnwidth]{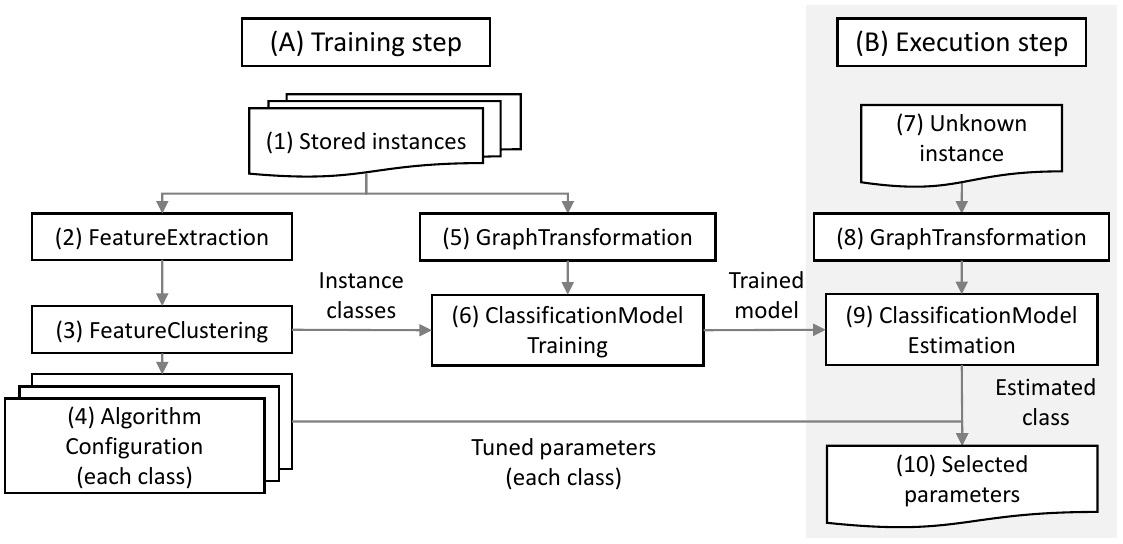}
        \caption{Overview of the proposed methodology framework.}
        \label{fig:Fig_fastISACwGNN}
    \end{minipage}
\end{figure}

Fig.~\ref{fig:Fig_fastISACwGNN} provides an overview of our proposed parameter selection framework.
This framework builds upon the ISAC method.
In our framework, instance clustering is based on their features, while the classification of new instances utilizes a graph structure.
This approach is adopted because the training step employs features that, although computationally intensive, have a direct impact on solver performance.
Conversely, during the execution step, the focus is on reducing computational time by selecting parameters quickly.
This contributes to minimizing the sum of the $TTS$ and the $T_{tune}$ in the execution step when the user solves an unknown instance.

During the training step, feature extraction and clustering are performed on the stored instances, as illustrated in the left section of Fig.~\ref{fig:Fig_fastISACwGNN}-(A).
Parameters for each cluster are then computed using AC method.
In the middle section, instances are converted into graph representations for input into a GNN, with their respective classes serving as ground truth labels.
Then, a GNN-based classification model to classify any instance accurately.
In the execution step, shown in Fig.~\ref{fig:Fig_fastISACwGNN}-(B), a new, unknown instance is converted into a graph representation.
The GNN then predicts its class, and the parameters associated with the predicted class, as determined by the AC method, are selected.

In this study, computationally intensive features extracted in the "FeatureExtraction" phase of Fig.~\ref{fig:Fig_fastISACwGNN}-(2), which are crucial for solver performance, include summary statistics from the history of feasible solutions and a set of feasible solutions obtained from actual solver executions.
Although feature extraction alone can take several minutes, "FeatureClustering" process enables us to group instances with similar solver search behaviors into clusters.
This is effective even when instances differ in task type, problem type, graph size, or density.
For the classification model, the matrix $\bm{Q}$ is treated as an adjacency matrix and converted into a graph representation using "GraphTransformation."

Additionally, table.~\ref{tb:symbol_list} lists the important symbols used in this paper and their descriptions.

\begin{table}[htbp]
    \centering
    \begin{tabular*}{1.0\linewidth}{@{\extracolsep{\fill}}clcl} \hline
        symbol & description & symbol & description \\ \hline
        $n$ & Number of binary variables. & $\Pi$ & Set of instances. \\
        $m$ & \parbox{0.3\linewidth}{Number of inequality constraints.} & $\pi$ & An instance. \\
        $\bm{x}$ & $n$-Dimensional binary vector. & $\bm{S}$ & Outputs of $\mathcal{T}$. \\
        $\bm{Q}$ & $n{\times} n$ real symmetric matrix. & $\bm{V}$ & Features converted from $\Pi$. \\
        $\bm{A}$ & $m{\times} n$ real matrix. & $\bm{G}$ & Graphs converted from $\Pi$. \\
        $\bm{b}$ & $m$-Dimensional real vector. & $g$ & Graph converted from $\pi$. \\
        $\mathcal{T}$ & Any algorithm. & $\tilde{g}$ & Graph sampled from $g$. \\
        $\theta$ & Parameter vector of $\mathcal{T}$. & $\Xi$ & GNN training data set. \\
        $\theta^*$ & \parbox{0.3\linewidth}{Optimized/estimated parameter vector.} & $\bm{C}$ & Set of classes of $\Pi$. \\
        $\Theta$ & Parameter spaces of $\mathcal{T}$. & $L$ & Number of classes. \\
        $F_{\mathcal{T}}$ & Objective function of $\mathcal{T}$. & $C^*$ & Estimated class. \\
        & & $\mathcal{M}_{GNN}$ & GNN model. \\
        \hline
    \end{tabular*}
    \caption{
        Symbols and their descriptions.
        }
    \label{tb:symbol_list}
\end{table}

\subsection{Heuristic Solver} \label{sec:heuristic_solver}

This section describes the heuristic solver and its parameters

In this study, DA~\cite{Fujitsu_DigitalAnnealer}, a software/hardware hybrid solver designed for solveing BQP problems, is used as the target solver.
The general processing flow of DA consists of three broad steps:
\begin{enumerate}[label=(\roman*)]
    \item Generation of random initial values.
    \item Solution search.
    \item Constraint satisfaction and convergence judgment.
\end{enumerate}
One cycle of this flow is called an epoch.

\subsection{Training Step} \label{sec:training_step}
\subsubsection{Feature Extraction} \label{sec:training_step_feature_extraction}

This section describes the "FeatureExtraction" component shown in Fig.~\ref{fig:Fig_fastISACwGNN}-(2) (Fig.~\ref{fig:Fig_fastISACwGNN_FeatureExtractionAndClustering} shows more details.).
"FeatureExtraction" takes an instance set $\Pi$ and run a heuristic solver $\mathcal{T}$ (DA in our case) and outputs a feature set $\bm{V}$.
The process involves extracting solver features from the instances and applying dimensionality reduction techniques.

The instance features utilize summary statistics from the update histories of feasible solutions and a feasible solution pool, which serve as logs of the DA execution results.
For the update histories, whenever the cost value (objective value) as defined in \eqref{eq:bqp_problem} is improved, the corresponding cost value and its timestamp are recorded.
In the feasible solution pool, solutions and their cost values are recorded or replaced every epoch, prioritizing solutions that satisfy constraints and have smaller cost values, up to the pool's capacity.
To mitigate the effects of randomness across various functions, DA is executed multiple times with different random seeds, and the logs of the results are concatenated.
The set of extracted features for each instance is denoted by $\bm{S}$.
While these features reflect the behavior of the solver, the extraction process takes several seconds to tens of seconds.

In order to improve a generalization ability of a machine learning model by removing noise and redundancy while preserving the important features of the data, dimensionality reduction is performed.
For this process, the uniform manifold approximation and projection (UMAP)~\cite{2018arXiv180203426M} method is employed.
UMAP assumes that high-dimensional data lie on a low-dimensional manifold and aims to find a low-dimensional representation that preserves the local and global structure of the data manifold.
Unlike principal component analysis, a common dimensionality reduction technique, UMAP effectively captures nonlinear data structures.
Furthermore, compared to t-distributed stochastic neighbor embedding~\cite{JMachLearnResearch2008V9p2579-2605}, UMAP has lower computational cost, making it advantageous for large-scale data, while simultaneously capturing both local and global data structures.

The UMAP algorithm is executed as follows:
First, for each point in the solver features $\bm{S}$ recorded for each instance, a neighborhood graph is constructed by connecting points that are close to each other in the high-dimensional space.
Next, the algorithm approximates the manifold to identify a low-dimensional representation that preserves the local and global structure of this neighborhood graph.
Finally, the data points are projected onto the low-dimensional space while maintaining the relationships optimized by the manifold approximation, resulting in the feature set $\bm{V}$ for all instances.

\subsubsection{Feature Clustering} \label{sec:training_step_feature_clustering}

In Fig.~\ref{fig:Fig_fastISACwGNN}-(3), the "FeatureClustering" component conducts clustering on the set of features $\bm{V}$ extracted by the "FeatureExtraction" process.
The hierarchical density-based spatial clustering of applications with noise (HDBSCAN)~\cite{ProcPAKDD2013p160-172} method is utilized for this clustering task.
HDBSCAN is a clustering technique that hierarchically partitions data based on local data density, making it appropriate for considering variations in data density.
Although g-means, used in the original ISAC, can automatically estimate the number of clusters, it assumes that a subset of the data follows a Gaussian distribution and is unsuitable for the irregular shape of the feature distribution.

The HDBSCAN algorithm is executed as follows:
First, a dendrogram is created by hierarchical clustering based on mutual reachability distance.
This dendrogram is then compressed into a cluster hierarchy by considering the minimum cluster size.
Finally, clusters are extracted from the compressed dendrogram using a stability scale.

The input features $\bm{V}$ are clustered into $L$ clusters $\bm{C}=\{C_i\}_{i=0}^{L-1}$ by HDBSCAN.
The instance set $\Pi$ is divided into instance sets for each cluster $\Pi^{(\bm{C})}=\{\Pi^{(C_i)}\}_{i=0}^{L-1}$.
Each cluster $C_i$ contains $n_{C_i}$ instances, and the set of instances belonging to a certain cluster $C_i$ is $\Pi^{(C_i)} = \{\pi_{j}^{(C_i)}\}_{j=0}^{n_{C_i}-1}$.

\subsubsection{Algorithm Configuration} \label{sec:training_step_algorithm_configuration}
\subsubsection*{SMBO with Tree-structured Parzen Estimator}

An SMBO algorithm is commonly used to solve AC problems.
This method aims to build a surrogate model to approximate the performance of an algorithm $\mathcal{T}$, utilizing methods such as GPs, TPE, and RF.

The surrogate model is defined using the expected improvement (EI) criterion.
Given a history $\mathcal{H}$ of observed algorithm $\mathcal{T}$ parameter and performance pairs $(\theta,\ y_F)$, where $y_F=F_{\mathcal{T}}(\theta)$, a model $\mathcal{M}_S$ is constructed.
The next evaluation point $\theta'$ is then proposed by minimizing a surrogate function $S(\theta,\ \mathcal{M}_S) \equiv - \mathrm{EI}(\theta)$ derived from this model.

In our study in Fig.~\ref{fig:Fig_fastISACwGNN}-(4) (Fig.~\ref{fig:Fig_fastISACwGNN_AlgorithmConfiguration}), the "AlgorithmConfiguration" process utilizes TPE~\cite{NIPS2011V24p2546-2554} as a surrogate model for the EI in the SMBO algorithm.
TPE calculates the EI as follow:
\begin{equation}
    \label{eq:expected_improvement_tpe}
    \mathrm{EI}(\theta) \propto \left( \gamma + \frac{g(\theta)}{l(\theta)} (1-\gamma) \right)^{-1},
\end{equation}
where $l(\theta)$ and $g(\theta)$ are distributions estimated using the Parzen estimator for the historical point sequence $y_F < y_F^*$ and $y_F \geq y_F^*$, respectively.
The $y_F^*$ is a reference value for $y_F$, set as the top $100\gamma\ \% \ (0<\gamma<1)$ of the observed $y_F$ values in $\mathcal{H}$.
In proposing the next exploration candidate $\theta'$, the point minimizing the surrogate function $S(\theta,\ \mathcal{M}_S)$, i.e., the point that minimizes the density ratio $\frac{g(\theta)}{l(\theta)}$ constructed from $\mathcal{H}$ in \eqref{eq:expected_improvement_tpe}, is proposed.

\subsubsection*{Instance Selection \& Objective Value} \label{sec:training_step_instance_selection_and_objective_value}

To enhance the generalization capability to instances beyond those used for tuning within each class, multiple instances are employed for parameter tuning within each class.
Empirical evidence suggests that focusing on more challenging instances within a class often results in enhanced performance across other instances.
A simple and practical measure of an instance's difficulty is the time of the last update when the solver is executed for a certain period~\cite{ACC24p98-103}.

As shown in Fig~\ref{fig:Fig_fastISACwGNN_AlgorithmConfiguration}, for each class, parameters $\theta$ of the algorithm $\mathcal{T}$ are tuned using the AC method with TPE on the selected multiple instances.
This process is executed for all $L$ classes, resulting in tuned parameters $\theta^{(C_i)}$ for each class $C_i$.

During parameter tuning, suggesting parameters and algorithm evaluations are performed iteratively.
The evaluation value of the algorithm is the sum of the cost value obtained when searching for an instance and the time when this value is updated.
It is crucial to normalize the cost value for each instance.
Without normalization, instances with higher rates of cost value change will be disproportionately emphasized, leading to parameter values biased toward these instances.
Since the best known solution (BKS) is typically unknown in real-world problems, normalization using BKS is not feasible.
Therefore, we consider both the cost value and the time taken to achieve it, and we define a normalization approach based on execution results using default parameters during feature extraction:
\begin{equation}
    \label{eq:normalized_objective_value}
    \frac{E_i^{(trial)} - \mathrm{mean}\big(E_i^{(default)}\big)}{\mathrm{std}\big(E_i^{(default)}\big)}
    + \alpha\ \frac{T_i^{(trial)} - T_{\mathrm{limit}}}{T_{\mathrm{limit}}},
\end{equation}
where $E_i^{(trial)}$ and $T_i^{(trial)}$ represent the minimum cost value and its corresponding time for instance $i$ in each trial, while $E_i^{(default)}$ denotes the cost value when instance $i$ is executed with default parameters.
$\alpha$ indicates the parameter for adjusting the contribution from time.
The first term is the normalized cost term.
During feature extraction, multiple seeds are employed to mitigate the impact of random variations, allowing for processing with the mean and standard deviation.
This process is expected to keep the normalized cost within the range of $-1$ to $1$.
The second term represents how quickly the solution was obtained.
Even if identical cost values are obtained, achieving the value sooner can enhance the parameter quality.

\subsubsection{Instance Classification with Graph Neural Network} \label{sec:training_step_instance_classification_with_gnn}

GNNs are utilized for instance classification owing to their exceptional ability to directly handle complex graph-structured data.
They offer robust representation learning capabilities and flexibility~\cite{NNLS2021V32p4-24}.
This section describes the instance transformation into graphs and the GNN model in Fig.~\ref{fig:Fig_fastISACwGNN}-(5) and -(6) (Fig.~\ref{fig:Fig_fastISACwGNN_GraphTransformation} and \ref{fig:Fig_fastISACwGNN_GNNTraining}).

\subsubsection*{Transform to Graph}

To input BQP problems into a GNN, we need to convert them into graphs.
The basic graph structure is defined by a pair $\mathcal{G}=(\mathcal{V},\ \mathcal{E})$, where $\mathcal{V}$ is a set of nodes and $\mathcal{E} \subset \{(u,\ v)\ |\ u,\ v \in \mathcal{V} \}$ is a set of edges.
In GNNs, feature vectors $\mathbf{x}_{u} \in \mathbb{R}^F,\ \mathbf{e}_{v,u} \in \mathbb{R}^D$ ($F,\ D$ are the dimensions of node and edge features) are typically added to each node and edge, respectively.
Additionally, ground truth or label information $y$ corresponding to the task is included.

Here, the instance set $\Pi^{(\bm{C})}$ divided for each class is converted into a graph data set $\bm{G}$ consisting of $n$ graphs, each with a graph structure and a correct class label for each instance.

\subsubsection*{Graph Classification with Graph Neural Network}

GNNs employ a message passing mechanism to gather information from the neighboring nodes of each node, updating their own state with this aggregated data.
By repeating this process, the input node features are transformed into embedded representations reflecting the information of the entire graph, enabling predictions based on the graph structure.

This paper utilizes a graph attention network (GAT)~\cite{2017arXiv171010903V} as a GNN model.
GAT introduces an attention mechanism that assigns varying weights to the connections between neighboring nodes during the feature update process.
This capability allows GAT to more effectively identify and capture important relationships and structural features among nodes.
The message passing function of GAT is defined as follows:
\begin{equation}
    \mathbf{h}_v^{(l)} = \sigma \bigg( \sum_{u\in \mathcal{N}(v)} \alpha_{vu}^{(l)} \mathbf{W}^{(l)} \mathbf{h}_v^{(l-1)} \bigg),
\end{equation}
where $\sigma (\cdot)$ is the sigmoid activation function, $\mathcal{N}(v)$ is the set of nodes adjacent to node $v$, $\mathbf{W}$ is the learning parameter of the model, and $\mathbf{h}_v^{(l)}$ is the intermediate representation of node $v$ at layer $l$.
Here, the first layer $\mathbf{h}_v^{(1)}$ of the intermediate representation is feature vector $\mathbf{x}_{v}$ of node $v$.
Attention weight $\alpha_{vu}^{(l)}$ represents the strength of the connection between node $v$ and its neighboring node $u$, and it is calculated as follows:
\begin{equation}
    \alpha_{vu}^{(l)} = \mathrm{softmax} \left( g\left( \mathbf{a}^T \left[\mathbf{W}^{(l)} \mathbf{h}_v^{(l-1)} \Big\Vert \mathbf{W}^{(l)} \mathbf{h}_u^{(l-1)} \right]\right) \right),
\end{equation}
where $g (\cdot)$ is the LeakyReLU activation function, $\mathbf{a}$ represents the learning parameter, and $\|$ denotes the concatenation operation.
The softmax function ensures that the sum of attention weights for all neighboring nodes of node $v$ equals 1.

A challenge in graph classification problems using GNNs is the variability in the number of nodes across different graphs.
Since the message passing function generates an embedding for each node, resulting in $n$ embeddings for the entire graph, it makes direct comparison between graphs difficult.
To address this issue, graph pooling is employed.
Graph pooling consolidates the multiple node embeddings into a single comprehensive representation for the entire graph, $\mathbf{z}' = \psi(\mathbf{z}_0,\cdots,\mathbf{z}_{n-1})$.
$\mathbf{z}_i$ is the node embedding, $n$ is a number of nodes entered into the pooling layer, and $\psi$ is the pooling function, with mean/max/sum pooling being representative examples.

The embedded representation $\mathbf{z}'$ obtained from the pooling layer is passed through a fully connected layer, such as a linear layer or MLP, to predict the label vector: $\hat{\mathbf{Y}}=\mathrm{FC}(\mathbf{z}')$.
The GNN model parameters can be trained to minimize the cross-entropy loss between the model output and the correct label.
The cross-entropy loss is defined as $\mathcal{L} = - \sum_{y \in \mathcal{Y}} \mathrm{Y}_y \log \hat{\mathrm{Y}}_y$, where $\mathcal{Y}$ denotes the label set, and $\mathrm{Y}_y$ and $\hat{\mathrm{Y}}_y$ are the elements of the correct label vector $\mathbf{Y}$ and the GNN prediction $\hat{\mathbf{Y}}$ using one-hot encoding, respectively.

The GNN model described here is trained using the graph data set $\bm{G}$ as explained in Section~\ref{sec:experimental_setup_gnn_model_training}, resulting in a trained model $\mathcal{M}_{GNN}$.

\subsection{Execution Step} \label{sec:run_step}

This section describes the execution step process illustrated in Fig.~\ref{fig:Fig_fastISACwGNN}-(B) (Fig.~\ref{fig:Fig_fastISACwGNN_GNNEvaluation}) utilized by users.

When an unknown instance $\pi_{unk}$, the trained GNN model $\mathcal{M}_{GNN}$, and the tuned parameters $\bm{\theta}$ are provided as inputs.
$\pi_{unk}$ is first transformed into a graph representation $g_{unk}$ by "GraphTransformation" in Sec.~\ref{sec:training_step_instance_classification_with_gnn}.
Inputting $g_{unk}$ to $\mathcal{M}_{GNN}$ in inference mode yields the predicted class $C^*$.
Finally, the execution step ends by selecting the parameters $\theta^*$ corresponding to class $C^*$.

\section{Experimental Setup} \label{sec:experimental_setup}

This chapter describes the detailed settings and evaluation methods of our proposed method described in the previous section.

The experimental set-up involves two different computing environments.
For tasks such as feature extraction and clustering detailed in Sec.~\ref{sec:experimental_setup_feature_extraction_and_clustering} (illustrated in Fig.~\ref{fig:Fig_fastISACwGNN}-(2) and -(3)), parameter tuning in Sec.~\ref{sec:experimental_setup_algorithm_configuration} (in Fig.~\ref{fig:Fig_fastISACwGNN}-(4)), and the evaluation of DA for each parameter in Sec.~\ref{sec:experimental_setup_evaluation}, the first environment consists of a Linux machine running CentOS 7.2, equipped with an Intel Xeon E5-2698 v4 (2.20 GHz) CPU, 512 GB of RAM, and an NVIDIA Tesla P100 GPU.
Conversely, for GNN training in Sec.~\ref{sec:experimental_setup_gnn_classification} (in Fig.~\ref{fig:Fig_fastISACwGNN}-(5) and -(6)) and the evaluation of GNN classification in Sec.~\ref{sec:experimental_setup_evaluation} (in Fig.~\ref{fig:Fig_fastISACwGNN}-(8) and -(9)), the second environment utilizes an Ubuntu 22.04 LTS OS, featuring an Intel Xeon Gold 6148 (2.40 GHz) CPU, 384 GB of RAM, and an NVIDIA Tesla V100 GPU.

\subsection{Dataset Instances}

\begin{figure}[htbp]
    \centering
    \begin{minipage}[b]{0.245\columnwidth}
        \centering
        \includegraphics[width=\columnwidth]{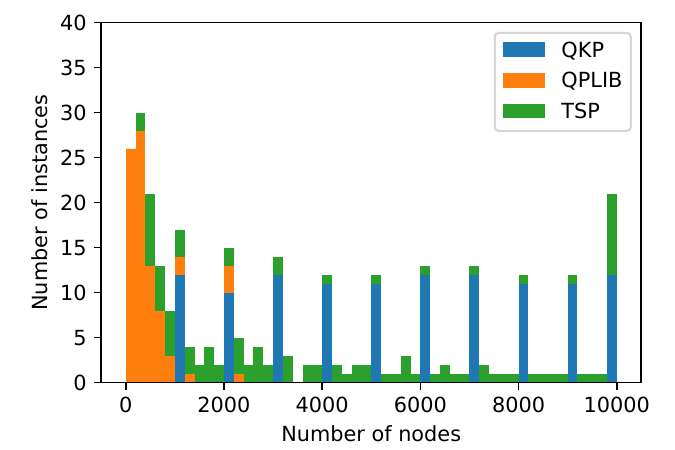}
        \subcaption{}
        \label{fig:fig_hist_Number_of_nodes}
    \end{minipage}
    \begin{minipage}[b]{0.245\columnwidth}
        \centering
        \includegraphics[width=\columnwidth]{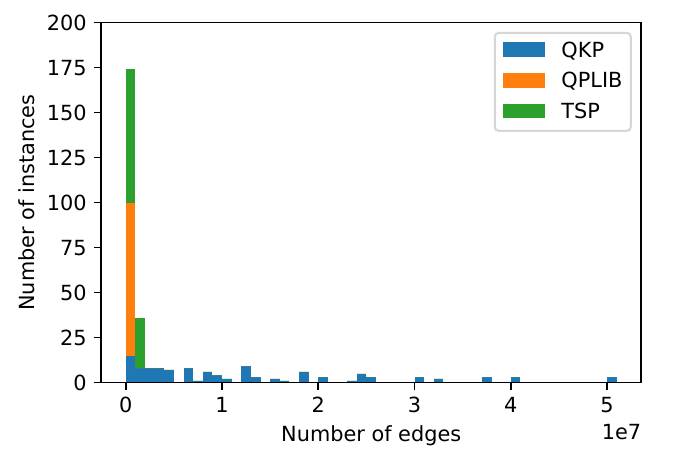}
        \subcaption{}
        \label{fig:fig_hist_Number_of_edges}
    \end{minipage}
    \begin{minipage}[b]{0.245\columnwidth}
        \centering
        \includegraphics[width=\columnwidth]{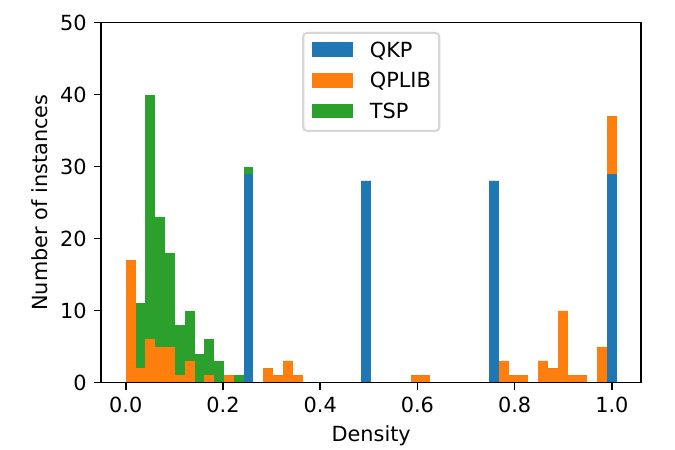}
        \subcaption{}
        \label{fig:fig_hist_Density}
    \end{minipage}
    \begin{minipage}[b]{0.245\columnwidth}
        \centering
        \includegraphics[width=\columnwidth]{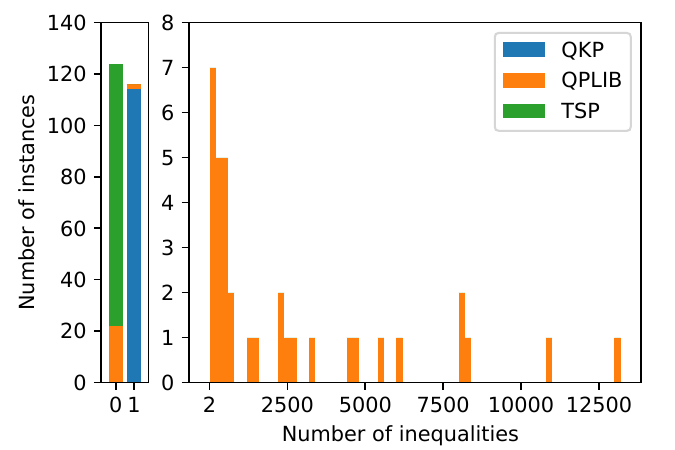}
        \subcaption{}
        \label{fig:fig_hist_Number_of_inequalities}
    \end{minipage}
    \caption{Distributions of graph statistics of the learning data set. (a) Number of nodes. (b) Number of Edges. (c) Density. (d) Number of inequality constraints.}
\end{figure}

The BQP problems utilized for training and evaluation data sets were prepared by integrating publicly available benchmark data with generated data.
For public data, we extracted instances from QPLIB~\cite{QPLIB2018}, which collects quadratic programming problems, TSPLIB~\cite{ORSAJComp1991V3p376-384}, which gathers TSP~\cite{fphy2014V2a5p1-15}, and MIPLIB~\cite{MIPLIB2017}, which assembles mixed-integer programming problems.
For generated data, we generated instances of TSP and quadratic knapsack problems (QKPs)~\cite{DiscApplMath2007V155p623-648}.
The BQPs in MIPLIB are linear problems, with the off-diagonal elements of the matrix $\bm{Q}$ being 0.
We assume that instances with a different distribution from QPLIB, TSP, and QKP are input, and use them only for evaluation.

The methods for extracting and generating instances for each benchmark data and problem class are as follows:

\begin{description}
    \item[QPLIB]~
    A well-known benchmark data set for quadratic programming problems.
    The selected instances include those with or without linear inequality constraints and feature binary variables numbering 10,000 or fewer, which are extracted and converted into BQPs.
    \item[TSPLIB]~
    A well-known benchmark data set for TSPs consisting of cities on a map.
    The instances used are those with 100 cities or fewer, extracted and converted into BQPs.
    \item[MIPLIB]~
    A well-known benchmark data set for mixed-integer programming problems.
    The instances used are randomly extracted from those with linear inequality constraints and binary variables of 10,000 or fewer, subsequently converting them into BQPs.
    \item[TSP (generated)]~
    Instances were generated with a range of 20 to 100 cities, incremented by 1.
    \item[QKP (generated)]~
    Instances were generated by combining the number of variables and densities.
    The number of variables ranges from 1,000 to 10,000, incremented by 1,000.
    The densities include values [0.25, 0.5, 0.75, 1.0].
    Three instances are generated from each combination of variable count and density.
\end{description}

After this, TSPLIB and TSP problem classes are collectively treated as TSP.
Excluding MIPLIB instances, the training data set comprises a total of $n=321$ instances.

Fig.~\ref{fig:fig_hist_Number_of_nodes} to \ref{fig:fig_hist_Number_of_inequalities} illustrate the distributions of variables, off-diagonal terms, density, and inequality constraints within the data set.
In terms of the number of variables, QPLIB instances are mostly below 1,000, TSP instances are uniformly distributed, and QKP instances are generated at every 1,000 increments.
In terms of density, QPLIB is divided into sparse and dense categories, TSP exhibits sparsity, and QKP is generated under conditions of 0.25, 0.50, 0.75, and 1.0 densities.
When examining inequality constraints, the data is divided into instances with 0 or 1 constraint and those with more.
TSP instances have no inequality constraints, while QKP instances have only one.
Conversely, QPLIB spans a range from instances without constraints to those with 10,000 or more.

\subsection{Heuristic solver} \label{sec:experimental_setup_heuristic_solver}

In the DA solver described in Sec.~\ref{sec:heuristic_solver}, three parameters play an important role in performance:
\begin{description}
    \item[\texttt{num\_run}]~
    Number of parallel executions.
    This parameter is crucial for generating solution diversity, as it involves executing runs with different initial values in parallel.
    \item[\texttt{gs\_level}]~
    A coefficient determining the maximum number of iterations (= number of variables $\times$ \texttt{gs\_level}) within one epoch.
    \item[\texttt{gs\_cutoff}]~
    It defines the number of iterations allowed within an epoch before terminating it if the minimum energy is not updated.
    This represents how quickly the search process can move away from a local minimum to start from a new initial value.
\end{description}
The parameter tuning in Sec.~\ref{sec:training_step_algorithm_configuration} is performed for these three parameters, and the GNNs in Sec.~\ref{sec:training_step_instance_classification_with_gnn} selects their values.

\subsection{Feature Extraction \& Clustering} \label{sec:experimental_setup_feature_extraction_and_clustering}

\begin{figure}[h]
    \centering
    \begin{minipage}[b]{0.245\columnwidth}
        \centering
        \includegraphics[width=\columnwidth]{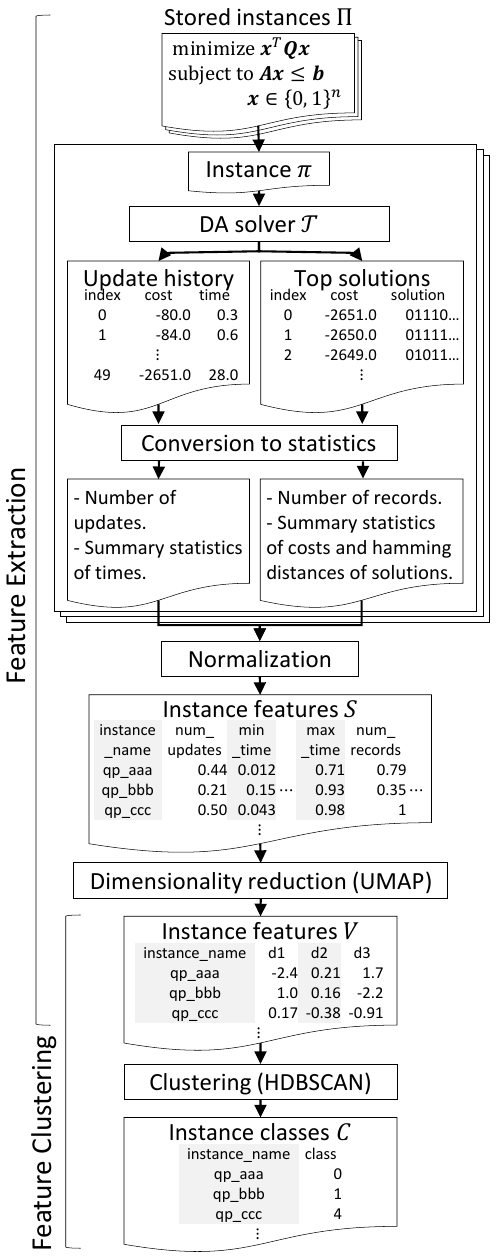}
        \subcaption{}
        \label{fig:Fig_fastISACwGNN_FeatureExtractionAndClustering}
    \end{minipage}
    \begin{minipage}[b]{0.245\columnwidth}
        \centering
        \includegraphics[width=\columnwidth]{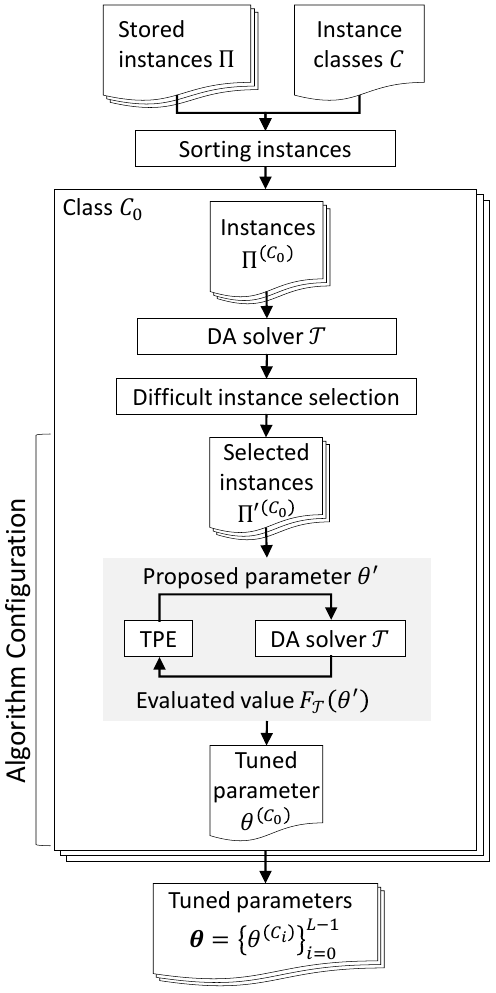}
        \vspace{0.5cm}
        \subcaption{}
        \label{fig:Fig_fastISACwGNN_AlgorithmConfiguration}
    \end{minipage}
    \begin{minipage}[b]{0.245\columnwidth}
        \centering
        \includegraphics[width=\columnwidth]{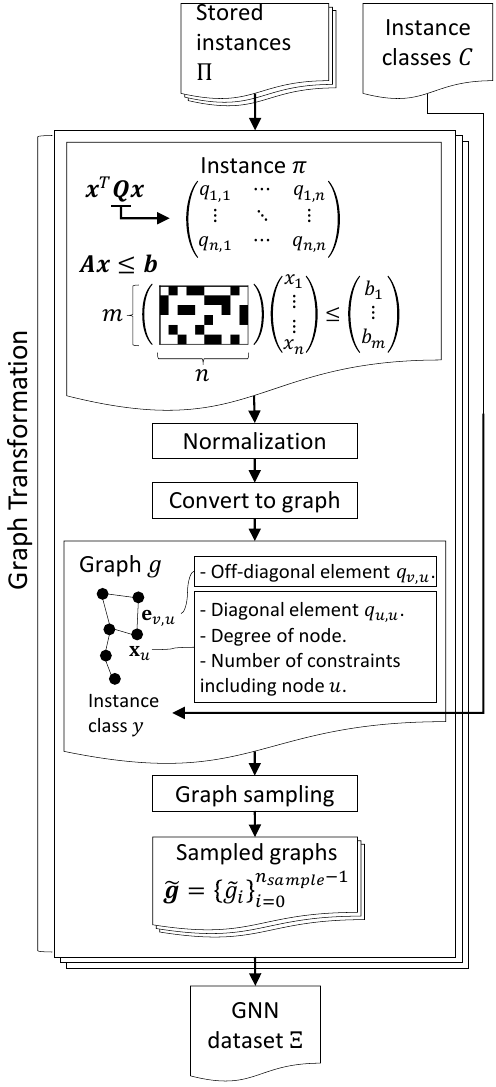}
        \vspace{0.1cm}
        \subcaption{}
        \label{fig:Fig_fastISACwGNN_GraphTransformation}
    \end{minipage}
    \begin{minipage}[b]{0.245\columnwidth}
        \centering
        \includegraphics[width=\columnwidth]{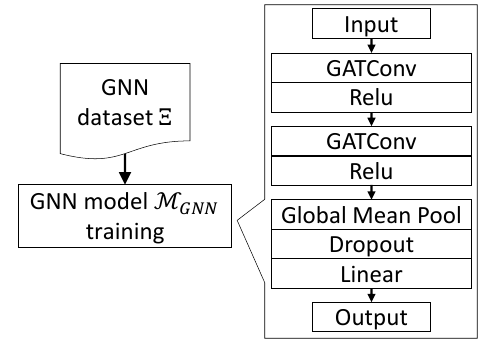}
        \vspace{0.2cm}
        \subcaption{}
        \label{fig:Fig_fastISACwGNN_GNNTraining}
        \vspace{1.0cm}
        \includegraphics[width=\columnwidth]{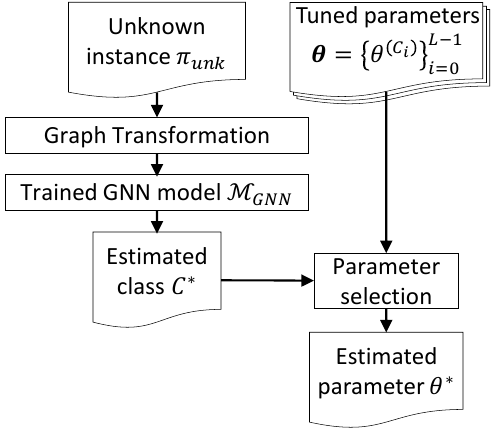}
        \vspace{0.2cm}
        \subcaption{}
        \label{fig:Fig_fastISACwGNN_GNNEvaluation}
    \end{minipage}
    \caption{Schematic diagrams of detail components in our framework. (a) Feature extraction and feature clustering. (b) Algorithm configuration. (c) Graph transformation. (d) GNN model training. (a) to (d) are in training step. (e) Execution step.}
\end{figure}

This section describes the processes of feature extraction, clustering, as outlined in Sec.~\ref{sec:training_step_feature_extraction} to \ref{sec:training_step_feature_clustering}.
The schematic diagrams of these processes are shown in Fig.~\ref{fig:Fig_fastISACwGNN_FeatureExtractionAndClustering}.

Features of an instance are derived using log information from the actual execution of the DA.
The DA operates with default parameters, including those intended for tuning in this study, and execution time is limited to 30 seconds.
The default values for the parameters targeted for tuning are listed in the "default" column of Table.~\ref{tb:hyperparameter}.
Log information includes a feasible solution pool, with up to 1,000 solutions, and an update history.
In addition, logs are gathered by running with 10 random number seeds to prevent the influence of seed numbers.
The total time of the actual DA exectution for feature extraction is $10\ \mathrm{seeds}\times 30\ \mathrm{s} = 300\ \mathrm{s}$.
The "FeatureExtraction" process in the execution step of the original ISAC method in Fig.~\ref{fig:Fig_ISAC}-(6) takes a similar amount of time and almost the entire time of the execution step in Fig.~\ref{fig:Fig_ISAC}-(B).

Information on the top solutions and the update history is used as summary statistics.
For the costs of top solutions, statistics such as the number of records, maximum, median, first and third quartiles of the cost ranking, which is sorted in ascending order based on cost values, are used.
Regarding the solutions for top solutions, the Hamming distance is calculated based on the solution with the minimum cost value, and its maximum, median, first and third quartiles are considered.
In addition, identical ratio defined by counting the number of variables inclinated to one side (0 or 1) in all solutions are calculated.
For the update history, the number of updates, minimum and maximum update times, and the median along with the first and third quartiles are used.
These statistics extracted for each instance are standardised and input to UMAP as instance features $\bm{S}$.

We used the \texttt{umap} package~\cite{JOSS2018V3p861} for UMAP for dimensionality reduction, as described in Sec.~\ref{sec:training_step_feature_extraction}.
There are four main parameters that can be adjusted in this package:
(i) \texttt{n\_neighbors}: The number of neighboring points to be recognized, which adjusts the balance between the local and global structure of the data.
(ii) \texttt{min\_dist}: The minimum distance among embedded points, which controls cluster density.
(iii) \texttt{n\_components}: The number of dimensions in the low-dimensional space.
(iv) \texttt{metric}: The method used to measure distance.

In this study, we set $\texttt{n\_neighbors}=30$ and $\texttt{min\_dist}=0.0$ based on the degree of feature aggregation, and $\texttt{n\_components}=3$ for effective visualization.
For $\texttt{metric}=\texttt{euclidean}$, the default value was used.

We used the \texttt{HDBSCAN} function of the \texttt{scikit-learn} package~\cite{JMLR2011V12p2825-2830} for HDBSCAN, a clustering technique described in Sec.~\ref{sec:training_step_feature_clustering}.
There are four main parameters that can be adjusted in this function:
(i) \texttt{min\_cluster\_size}: The minimum number of points required to form a cluster.
(ii) \texttt{min\_samples}: The number of neighboring points to define a cluster's core.
(iii) \texttt{cluster\_selection\_epsilon}: The distance between points to be grouped as a cluster.
(iv) \texttt{metric}: The method used to measure distance.

In this study, we set $\texttt{min\_cluster\_size}=30$ and $\texttt{min\_samples}=5$ to extract relatively large clusters for parameter tuning for each class using multiple instances.
$\texttt{cluster\_selection\_epsilon}=0.0$ and $\texttt{metric}=\texttt{euclidean}$ are aligned with $\texttt{min\_dist}=0.0$ and $\texttt{metric}=\texttt{euclidean}$ set in UMAP.

\subsection{Algorithm Configuration} \label{sec:experimental_setup_algorithm_configuration}

The parameter tuning for each class are executed using the method outlined in Sec.~\ref{sec:training_step_algorithm_configuration}.
The DA parameters $\theta$ to be tuned this time are introduced in Sec.~\ref{sec:heuristic_solver}.
The schematic diagram of tnuning parameters is shown in Fig.~\ref{fig:Fig_fastISACwGNN_AlgorithmConfiguration}.
Table.~\ref{tb:hyperparameter} provides a comprehensive overview of parameter types, default values, search ranges, and specifies whether sampling is performed in a logarithmically transformed space.

\begin{table}[htbp]
    \centering
    \begin{tabular*}{0.9\linewidth}{@{\extracolsep{\fill}}rrrrr} \hline
                   & type & default &       range & log scale \\ \hline
        num\_run   &  int &   16 &  $[1, 100]$ & False \\
        gs\_level  &  int &    5 &  $[1, 100]$ & False \\
        gs\_cutoff &  int & 8000 & $[1, 10^4]$ & True \\ \hline
    \end{tabular*}
    \caption{
        Tuning parameters, their types, default values, and search range.
        }
    \label{tb:hyperparameter}
\end{table}

Ten instances were used for tuning for each class based on the following conditions; each instance was executed with the DA for 60 seconds using default parameters, and those with higher difficulty, as determined by the latest solution update time, were selected.

In this experiment, we used Optuna~\cite{ProcKDD2019p2623-2631}, which implements TPE as a surrogate model for predicting search performance.
The tuning process involved Bayesian optimization with a total of 500 trials.
For the first 100 trials, random searches are executed to mitigate the influence of initial values.
The history of past trials was utilized for up to 24.
In addition, $\alpha=0.001$ is set as the coefficient for balancing the first and second terms in the cost value of \eqref{eq:normalized_objective_value} to be evaluated in each trial.

\subsection{GNN Classification} \label{sec:experimental_setup_gnn_classification}

The input to the GNN consists of a set of nodes, edges, their respective features, and the target value for the task as described in Sec.~\ref{sec:training_step_instance_classification_with_gnn}.
The schematic diagram of the graph transformation is shown in Fig.~\ref{fig:Fig_fastISACwGNN_GraphTransformation}.
Each node corresponds to a variable in the BQP, with features represented by a 3D vector.
This vector includes the value of the diagonal elements of matrix $\bm{Q}$, the degree representing the number of connected edges, and a number of times specified by the inequality constraints.
The node feature of the diagonal elements are normalized by the hyperbolic tangent after dividing by the absolute value of their own median.
The edges are non-zero elements of the off-diagonal elements of matrix $\bm{Q}$, representing connections among variables, with features set to the values of these non-zero elements.
The edge feature is normalized in the same manner as node feature of the diagonal elements.
The correct answer value for the task is the classification class number.

Node sampling is applied to the graph address GPU memory to address limitations and enhance data augmentation owing to limited data availability.
This involves randomly extracting $10\ \%$ of the nodes in the graph for each instance.
Edges are preserved only if both connected nodes remain post-sampling.
The classification class number of the original graph is assigned as the correct answer value for the task.
This operation is performed 20 times.

The overall structure of the GNN is shown in Fig.~\ref{fig:Fig_fastISACwGNN_GNNTraining}.
It consists of two layers of \texttt{GATConv} and \texttt{Relu}, followed by \texttt{global\_mean\_pool}, \texttt{dropout}, and \texttt{linear} layers.
\texttt{GATConv} specifies 3 node features, 8 hidden layer channels, 1 edge feature, 4 attention heads, self-loop enabled, and a dropout rate of 0.3.
Owing to varying sizes of graph inputs, a global mean pooling layer is applied after the GATConv layers
\texttt{dropout} is set to 0.3.
The output is prepared with a channel for each class in the \texttt{linear} layer.
The total number of parameters is about 600.

\subsubsection{GNN Model Training} \label{sec:experimental_setup_gnn_model_training}

During training, cross-entropy loss, as described in Sec.~\ref{sec:training_step_instance_classification_with_gnn}, is used as the criterion, while the Adam optimizer is employed with a default value of $\texttt{learning\_rate}=0.01$.
Training spans 500 epochs, allowing for sufficient convergence of the learning curve, and uses a batch size of 32 to prevent GPU memory overflow.
To avoid class imbalance in the data set, additional graphs from other classes are randomly selected, ensuring the total number of graphs is up to 1.2 times that of the smallest class.
This selected data set is divided into training and validation sets in a 9:1 ratio.
The model achieving the highest accuracy on the validation data is chosen as the final model.

\subsection{Evaluation} \label{sec:experimental_setup_evaluation}

To evaluate the class $C^*$ estimated by GNN and its parameters $\theta^*$, we compared the classification accuracy of GNN with the solver performance using selected parameters.
Table.~\ref{tb:instance_stats} lists the evaluation instances.
The "Instance ID" column denotes the ID for each instance, while the "problem" column specifies the type of problem each instance represents.
The "s" column indicates generated data with a $\checkmark$ for generated data.
The "n" column lists the number of variables, "m" details the number of inequality constraints, and "d" describes the graph density of the undirected graph, defined as $d=\frac{2|\mathcal{E}|}{|\mathcal{V}|*(|\mathcal{V}|-1)}$.
The instances from QPLIB, TSP, and QKP were pre-selected, totaling 20, to prevent any influence on feature generation in UMAP.
In addition, five instances were randomly extracted from the MIPLIB library, serving as instances outside the training data set distribution, and were only used to evaluate solver performance

The classification accuracy of the GNN model is evaluated using several metrics such as accuracy, precision, recall, F-value for each class, macro average, and confusion matrix.
The flow of classification by the trained GNN model is shown in Fig~\ref{fig:Fig_fastISACwGNN_GNNEvaluation}.
The node sampling of each instance is conducted 20 times at a $10\ \%$ rate, similar to the training data process outlined in Sec.~\ref{sec:experimental_setup_gnn_classification}.
The classification result is determined by the most frequently occurring class.

As a result of the classification, the performance of the selected parameters is evaluated relative to (I) the default parameters of the DA, and (II) the cost value and time required to achieve a feasible solution using parameters derived from 10 instances for each problem type, along with the tuning duration.
The cost value and time to reach that value are averaged from results obtained by running the DA for 30 seconds using 10 different seeds.
The tuning time $T_{tune}$ is the sum of the results of inferring the node-sampled graph 20 times with the GNN, excluding the disk IO processing time.
Cost value comparisons with (I) and (II) use the gap defined between the cost value with selected parameters $E^{(class)}$ and the default result $E^{(default)}$ or the result tuned for each problem $E^{(problem)}$.

\begin{equation}
    \label{eq:gap}
    \mathrm{Gap} = 100 \times \frac{E^{(class)} - E^{(\#)}}{|E^{(\#)}|}\quad (\#: \mathrm{default/problem}).
\end{equation}
The comparison of the time to reach the final cost value involves evaluating the difference in $TTS$ for each parameter, $\Delta TTS = TTS^{(class)} - TTS^{(\#)}$.
Since class classification requires tuning time $T_{tune}$, the following is used for the overall time comparison:
\begin{equation}
    \label{eq:total_time}
    \Delta T_{tot} = \Delta TTS + T_{tune}.
\end{equation}

\section{Result \& Discussion} \label{sec:result_and_discussion}
\subsection{Feature Extraction \& Clustering} \label{sec:res_disc_feat_ext_and_clust}

\begin{figure}[htbp]
    \centering
    \begin{minipage}[b]{0.49\columnwidth}
        \centering
        \includegraphics[width=0.75\columnwidth]{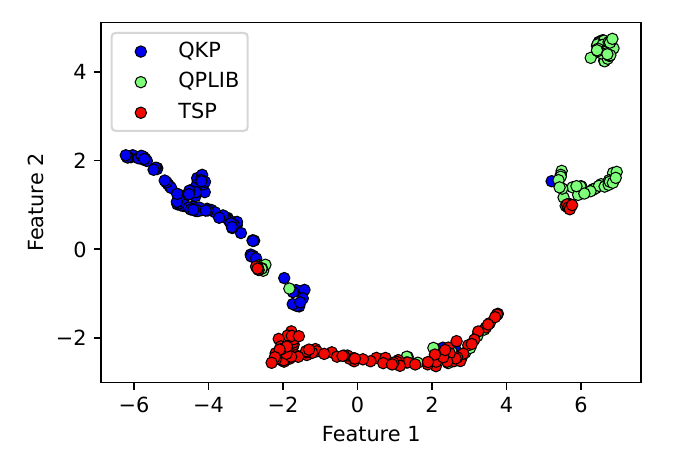}
        \subcaption{}
        \label{fig:fig_scatter}
    \end{minipage}
    \begin{minipage}[b]{0.49\columnwidth}
        \centering
        \includegraphics[width=0.75\columnwidth]{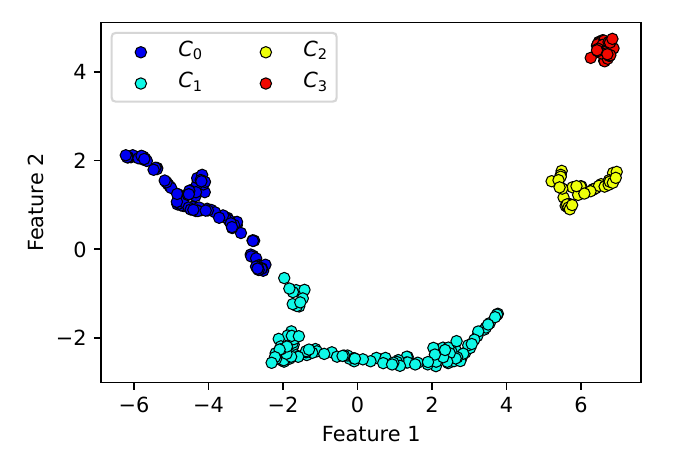}
        \subcaption{}
        \label{fig:fig_scatter_cluster}
    \end{minipage}
    \caption{(a) Instance distributions colored by problem in the feature space. (b) Instance distributions colored by clustering results in the feature space.}
\end{figure}

Fig.~\ref{fig:fig_scatter} and \ref{fig:fig_scatter_cluster} illustrate the results of color-coding features extracted from instances using the method outlined in Sec.~\ref{sec:training_step_feature_extraction} for each problem type, as well as the results from clustering using the method described in Sec.~\ref{sec:training_step_feature_clustering}.
Table.~\ref{tb:prj_cls_instances} provides a breakdown of the number of instances sorted by problem type and clustering category.

In Fig.~\ref{fig:fig_scatter}, it is evident that instances are distinctly clustered in the feature space.
Although clusters are roughly divided by problem type, some instances appear within clusters of different problem types, indicating that even within the same problem category, there are instances exhibiting search behavior similar to other problem types.

Fig.~\ref{fig:fig_scatter_cluster} shows that the instances are divided into four classes: $\bm{C}=\{C_i\}_{i=0}^3$, based on the structure in the feature space.
Table.~\ref{tb:prj_cls_instances} illustrates that QKP and TSP tend to be grouped into classes $C_0$ and $C_1$, respectively, while QPLIB, which generally handles quadratic programming problems, tends to be distributed among multiple classes: $C_1$, $C_2$, and $C_3$.

\begin{table}[htbp]
    \centering
    \begin{tabular*}{0.9\linewidth}{@{\extracolsep{\fill}}rrrrrr} \hline
                  & \C{0} & \C{1} & \C{2} & \C{3} & Total \\ \hline
        \C{QKP}   &    96 &    12 &     1 &     5 &   114 \\
        \C{QPLIB} &     6 &    13 &    37 &    29 &    85 \\
        \C{TSP}   &     6 &    90 &     6 &     0 &   102 \\ \hline
            Total &   108 &   115 &    44 &    34 &   301 \\ \hline
    \end{tabular*}
    \caption{
        Issue and class instance breakdown.
        }
    \label{tb:prj_cls_instances}
\end{table}

\subsection{Algorithm Configuration}

Table.~\ref{tb:selected_instances} presents a detailed breakdown of instances selected for each problem type using the instance selection method described in Sec.~\ref{sec:training_step_instance_selection_and_objective_value}.
Selection categories \C{0}, \C{1}, and \C{2} reflect the distribution of problems in the training data, as shown in Table.~\ref{tb:prj_cls_instances}.
On the other hand, in $C_3$, all five QKP instances in that class are selected.

Table.~\ref{tb:hyp_opt_class_prject} outlines the default parameters of the DA, the parameters calculated for each problem, and the results calculated for each class.

The parameter \texttt{num\_run} for \C{QKP} is adjusted to a smaller value compared to the default.
This indicates a preference for executing each \texttt{run} at high speed to quickly achieve a better solution, even if this reduces the degree of parallelism and, consequently, the diversity of solutions during the search.
This pattern is also evident in \C{0}, which consists of QKP instances, and \C{MIPLIB}, which is introduced only for evaluation.
Conversely, parameter \texttt{num\_run} for \C{QPLIB} and \C{TSP} is increased compared to the default, suggesting that a greater diversity of solutions is beneficial during the search process.

Parameter \texttt{gs\_level}, which determines the length of one \texttt{run}, is larger for all tuned parameters compared to the default value.
Each \texttt{run} involves variable inversion over $n \times \texttt{gs\_level}$ iterations to minimize cost values.
This implies that extending the duration of a single \texttt{run} is advantageous for achieving optimal results across all classes and problems.

Parameter \texttt{gs\_cutoff}, which monitors the history of the cost values of each \texttt{run} and terminates the search if the cost value has not been updated for a certain number of iterations.
For the tuned parameters, excluding \C{MIPLIB}, \texttt{gs\_cutoff} is set lower than the default, likely due to the smaller instance sizes used in this experiment, which involve 10,000 variables or fewer, as depicted in Fig.~\ref{fig:fig_hist_Number_of_nodes}.
The default value of parameter $\texttt{gs\_cutoff}=8000$.
QKP and \C{0}, which consist of QKP instances, have relatively low values, around several tens, while TSP and \C{1}, which mainly consist of TSP, have values of around several hundreds, \C{MIPLIB} shows larger values of around several thousands.
This difference reflects that some configurations opt to move to the next epoch quickly as soon as finding a local solution to escape the local solution, while others wait longer to overcome the local solution.
In addition, while \C{QPLIB} has very small values of $\texttt{gs\_cutoff}=1$, $C_2$ and $C_3$, which contain many QPLIBs, have larger values, indicating that both of the above strategies are effective.

\begin{table}[htbp]
    \centering
    \begin{tabular*}{0.9\linewidth}{@{\extracolsep{\fill}}rrrrrr} \hline
        & \C{0} & \C{1} & \C{2} & \C{3} \\ \hline
    \C{QKP}   &    10 &     0 &     1 &     5 \\
    \C{QPLIB} &     0 &     2 &     8 &     5 \\
    \C{TSP}   &     0 &     8 &     1 &     0 \\ \hline
    \end{tabular*}
    \caption{
        Breakdown of problem types for instances selected to fine-tune parameters by class.
        }
    \label{tb:selected_instances}
\end{table}

\begin{table*}[htbp]
    \centering
    \begin{tabular*}{0.9\linewidth}{@{\extracolsep{\fill}}r|r|rrrr|rrr:r} \hline
                   & default & \C{0} & \C{1} & \C{2} & \C{3} & \C{QKP} & \C{QPLIB} & \C{TSP} & \C{MIPLIB} \\ \hline
        num\_run   &      16 &     3 &    64 &    55 &    58 &       3 &        23 &      62 &    1 \\
        gs\_level  &       5 &    22 &    45 &    65 &    67 &      77 &        59 &      17 &   98 \\
        gs\_cutoff &    8000 &    61 &   845 &   255 &   167 &      23 &         1 &     571 & 9954 \\ \hline
    \end{tabular*}
    \caption{
        Parameter tuning results for each class/problem.
        }
    \label{tb:hyp_opt_class_prject}
\end{table*}

\subsection{GNN Classification} \label{sec:res_disc_gnn_class}

This section describes the instance sampling results and the prediction results of the test data using GNNs.

Table.~\ref{tb:gnn_dataset} shows the breakdown of training, validation, and test data obtained by sampling the graphed instances, as described in Sec.~\ref{sec:experimental_setup_gnn_classification}.
To minimize bias in each class, training and validation data are randomly sampled from $C_0$ to $C_2$, ensuring that their numbers do not exceed 1.2 times that of \C{3}, the class with the smallest number of instances.

Table.~\ref{tb:test_dataset_metrics} presents the GNN's prediction results for the test data, while the second column of Table.~\ref{tb:cost_gap_last_update_time_diff_default} lists the true versus predicted values for each class.
The true class was determined by extracting features from the evaluation instances, superimposing them on Fig.~\ref{fig:fig_scatter_cluster}, and using the nearest cluster as the true class.
For predicting test data classes, which are also sampled, a majority vote among instances from the same sampling source is conducted, and this outcome serves as the predicted class.

The recall data in Table.~\ref{tb:test_dataset_metrics} highlights instances of misclassification within true classes \C{1} and \C{2}.
QKP\_GEN\_004, which was misclassified as true class \C{1}, was classified into \C{0}, which is mainly composed of the same type of QKP.
Similarly, QPLIB\_10048 and QPLIB\_10061, intended to be classified as true class \C{2}, were incorrectly classified into \C{3}, which is half composed of the same type of QPLIB.
These errors underscore the challenges faced by the GNN in capturing the latent representation of graph structures, particularly concerning the solver behavior.

\begin{table}[htbp]
    \centering
    \begin{tabular*}{0.9\linewidth}{@{\extracolsep{\fill}}rrrrrr} \hline
                   & \C{0} & \C{1} & \C{2} & \C{3} & total \\ \hline
        training   & 734 & 734 & 734 & 612 & 2814 \\
        validation &  82 &  82 &  82 &  68 &  314 \\
        test       & 100 & 100 & 100 & 100 &  400 \\ \hline
    \end{tabular*}
    \caption{
        Breakdown of data set used for GNN learning.
        }
    \label{tb:gnn_dataset}
\end{table}

\begin{table}[htbp]
    \centering
    \begin{tabular*}{0.9\linewidth}{@{\extracolsep{\fill}}rrrr} \hline
           & precision & recall & f1-score \\ \hline
        \C{0} & 0.83 & 1.00 & 0.91 \\
        \C{1} & 1.00 & 0.80 & 0.89 \\
        \C{2} & 1.00 & 0.60 & 0.75 \\
        \C{3} & 0.71 & 1.00 & 0.83 \\
        macro avg & 0.89 & 0.85 & 0.85 \\ \hline
        \multicolumn{3}{c}{accuracy} & 0.85 \\ \hline
    \end{tabular*}
    \caption{
        Accuracy evaluation results of trained GNN model with evaluation data set.
        }
    \label{tb:test_dataset_metrics}
\end{table}

\begin{table}[htbp]
    \centering
    \begin{tabular*}{0.9\linewidth}{@{\extracolsep{\fill}}rrrrrr} \hline
        &       & \multicolumn{4}{c}{Predict} \\
        &       & \C{0} & \C{1} & \C{2} & \C{3} \\ \hline
        \multirow{4}{*}{True} & \C{0} & 5 & 0 & 0 & 0 \\
        & \C{1} & 1 & 4 & 0 & 0 \\
        & \C{2} & 0 & 0 & 3 & 2 \\
        & \C{3} & 0 & 0 & 0 & 5 \\ \hline
    \end{tabular*}
    \caption{
        Confusion matrix with an evaluation data set of trained GNN model.
        }
    \label{tb:confusion_matrix}
\end{table}

\subsection{DA Evaluation}

Table.~\ref{tb:cost_gap_last_update_time_diff_default} presents the outcomes of executing the DA on test data using parameters selected through class classification as detailed in Sec.~\ref{sec:res_disc_gnn_class} (columns 3 to 6 of Table.~\ref{tb:hyp_opt_class_prject}), alongside results obtained with the default parameters (column 2 of Table.~\ref{tb:hyp_opt_class_prject}).
Table.~\ref{tb:cost_gap_last_update_time_diff_problem} compares the DA performance on test data using parameters specifically tuned for each problem (columns 7 to 10 of Table.~\ref{tb:hyp_opt_class_prject}) against those selected by class classification.
The "Instance ID" column identifies the instances evaluated, specifying the problem type, whether the instance is generated, and including a unique string.
Generated instances are denoted with the prefix \texttt{GEN} and are sequentially numbered starting from 000.
The name of the library is used for instances from the library.
The "GNN($C_X$)" column displays the results of GNN class classification, as described in the previous Sec.~\ref{sec:res_disc_gnn_class}.
MIPLIB instances, introduced as out-of-distribution data, do not have true values listed; instead, they are represented by a horizontal line.
The "Gap" column shows the gap defined by \eqref{eq:gap} between the cost value of the result of class classification $E^{(class)}$ and the cost value of the comparison target $E^{(\#)}\ (\#= \mathrm{default/problem})$.
The "$\Delta TTS$" column indicates the time difference to reach the final cost when running with each parameter $TTS^{(class)} - TTS^{(\#)}$, the "$T_{tune}$" column shows the processing time for class classification by GNN excluding disk IO, while the "$\Delta T_{tot}$" presents the total time difference from default or problem-based results, which is the sum of $\Delta TTS$ and $T_{tune}$ defined by \eqref{eq:total_time}.
Bold values in the "Gap" and "$\Delta T_{tot}$" columns indicate significant changes, with improvements marked as $-1\ \%$ or less, and $-1$ second or less for "$T_{tot}$."
Reflects user perception of performance, marked with $\bigcirc\ /\ \times$.
$\bigcirc$ for improvements of $-1\ \%$ or less in the "Gap" column and $-1$ second or less in the "$\Delta T_{tot}$" column, and $\times$ is marked for those with an improvement of more than $-1\ \%$ in the "Gap" column and more than 1 second in the "$\Delta T_{tot}$" column.
Unmarked entries are deemed to have equivalent performance to the default parameters.

In the comparison with the execution results using the default parameters in Table.~\ref{tb:cost_gap_last_update_time_diff_default}, the "Gap" and "$\Delta TTS$" columns improvements reveal notable improvements in both cost and execution speed.
Execution speed enhancements are particularly evident across many instances.
When evaluating overall execution time, the GNN processing time shown in the "$T_{tune}$" column is $< 0.3\ \mathrm{s}$, allowing the overhead of $T_{tune}$ to be negligible for $\Delta TTS$ values ranging from several seconds to tens of seconds, thus resulting in a beneficial speedup.
On the other hand, for $\Delta TTS$ of less than 1 second, the time taken for $T_{tune}$ can lead to slower execution speeds.
Improvements are observed in $\Delta TTS$ for QKP and MIPLIB, and in both Gap and $\Delta TTS$ for TSP.
For QPLIB, as shown in the "n" column of Table.~\ref{tb:instance_stats}, the instance size is small, and the instances are easy to search even with the default parameters, so no substantial differences in Gap and $\Delta TTS$ are noted.

In comparing execution results using parameters tuned for each problem in Table.~\ref{tb:cost_gap_last_update_time_diff_problem}, improvements and deteriorations vary across instances.
Although $\Delta TTS$ shows mixed results, there is minimal difference in Gap, remaining below $1\ \%$, indicating that class classification has a limited impact on QKP and TSP owing to their constrained instance distributions.
For QPLIB, a significant difference is observed in $\Delta TTS$ for instances QPLIB\_3832 and QPLIB\_6647, where the true class \C{2} was correctly predicted, indicating the benefits of dividing QPLIB into multiple classes.
However, as no difference in $\Delta TTS$ is observed for these instances in Table.~\ref{tb:cost_gap_last_update_time_diff_default}, it suggests that QPLIB-tuned results are targeted at a narrower range of problems, and default parameters maintain some generalizability.
In addition, improvements are observed in both Gap and $\Delta TTS$ for MIPLIB, and as in the case of QPLIB, DA performance is improved by covering MIPLIB with parameters for multiple classes.
In particular, for MIPLIB\_neos-2328163-agri, a feasible solution is obtained using class-classified parameters, whereas no feasible solution was found in any of the 10 trials with parameters tuned for MIPLIB instances.

As an overall evaluation, counting $\bigcirc\ /\ \times$ in the "Effect" column, the comparison with default parameters shows 14 instances as effective, 11 as equivalent, and none as worse.
When compared with problem-specific parameters, there are 8 effective, 15 equivalent, and 2 worse instances.

This analysis suggests that the method significantly improves search performance compared to default parameters.
It also achieves nearly the same performance level as parameters tailored by problem type classification, which were not initially known.

{
    \tabcolsep=2pt

    \begin{table}[htbp]
        \centering\scriptsize
        \begin{tabular}{|l|r|c|r|r|r|} \hline
                Instance ID & Problem & s & n & m & d \\ \hline
            QKP\_GEN\_000             & QKP    & $\checkmark$ & 9000 &     1 & $        0.250$ \\
            QKP\_GEN\_001             & QKP    & $\checkmark$ & 2000 &     1 & $        0.500$ \\
            QKP\_GEN\_002             & QKP    & $\checkmark$ & 2000 &     1 & $        0.500$ \\
            QKP\_GEN\_003             & QKP    & $\checkmark$ & 4000 &     1 & $        0.750$ \\
            QKP\_GEN\_004             & QKP    & $\checkmark$ & 5000 &     1 & $        0.750$ \\
            QKP\_GEN\_005             & QKP    & $\checkmark$ & 8000 &     1 & $        1.000$ \\
            QPLIB\_0633               & QPLIB  &              &   75 &     2 & $        1.000$ \\
            QPLIB\_3750               & QPLIB  &              &  210 &   140 & $        0.330$ \\
            QPLIB\_3832               & QPLIB  &              &  561 &     0 & $        0.007$ \\
            QPLIB\_6647               & QPLIB  &              &  627 &    66 & $        0.072$ \\
            QPLIB\_10048              & QPLIB  &              &  150 &     5 & $        0.986$ \\
            QPLIB\_10058              & QPLIB  &              &  200 &    11 & $        0.879$ \\
            QPLIB\_10061              & QPLIB  &              &  200 &     5 & $        0.899$ \\
            QPLIB\_10065              & QPLIB  &              &  200 &    11 & $        0.970$ \\
            QPLIB\_10067              & QPLIB  &              &  200 &     5 & $        0.987$ \\
            TSP\_GEN\_000             & TSP    & $\checkmark$ & 1600 &     0 & $        0.099$ \\
            TSP\_GEN\_001             & TSP    & $\checkmark$ & 3481 &     0 & $        0.067$ \\
            TSP\_GEN\_002             & TSP    & $\checkmark$ & 5041 &     0 & $        0.056$ \\
            TSPLIB\_burma14           & TSP    &              &  196 &     0 & $        0.277$ \\
            TSPLIB\_hk48              & TSP    &              & 2304 &     0 & $        0.083$ \\
            MIPLIB\_cvs16r128-89      & MIPLIB &              & 3472 &  4633 & $            0$ \\
            MIPLIB\_neos-2328163-agri & MIPLIB &              & 2236 &  1963 & $            0$ \\
            MIPLIB\_seymour           & MIPLIB &              & 1372 &  4944 & $            0$ \\
            MIPLIB\_queens-30         & MIPLIB &              &  900 &   960 & $            0$ \\
            MIPLIB\_reblock354        & MIPLIB &              & 3540 & 19906 & $            0$ \\ \hline
        \end{tabular}
        \caption{
            Instance information used for evaluation.
            The "Instance ID" column is the instance ID, the "problem" column represents the instance problem type, and the "s" columns contain artificially generated data with a $\checkmark$.
            The "n" column denotes the number of variables, the "m" column indicates the number of inequality constraints, and the "d" column is the undirected graph density $d=\frac{2\left|\mathcal{E}\right|}{\left|\mathcal{V}\right|*\left(\left|\mathcal{V}\right|-1\right)}$.
            }
        \label{tb:instance_stats}
    \end{table}
}

{
    \tabcolsep=2pt

    \begin{table}[htbp]
        \centering\scriptsize
        \begin{tabular}{|l|c|r|r|r|r|c|} \hline
            \multirow{2}{*}{Instance ID} & GNN(\C{X}) & \multicolumn{1}{|c}{Gap} & \multicolumn{1}{|c}{$\Delta TTS$} & \multicolumn{1}{|c}{$T_{tune}$} & \multicolumn{1}{|c|}{$\Delta T_{tot}$} & \multirow{2}{*}{Effect} \\
            & True/Pred & \multicolumn{1}{|c}{[\%]} & \multicolumn{1}{|c}{[s]} & \multicolumn{1}{|c}{[s]} & \multicolumn{1}{|c|}{[s]} & \\ \hline
            QKP\_GEN\_000             & 0 / 0 & $    \bm{-23.3}$ & $         -3.16$ & $        0.211$ & $    \bm{-2.95}$ & $\bigcirc$ \\
            QKP\_GEN\_001             & 0 / 0 & $        -0.025$ & $         -25.8$ & $        0.145$ & $    \bm{-25.6}$ & $\bigcirc$ \\
            QKP\_GEN\_002             & 0 / 0 & $        -0.045$ & $         -27.9$ & $        0.169$ & $    \bm{-27.7}$ & $\bigcirc$ \\
            QKP\_GEN\_003             & 0 / 0 & $        -0.623$ & $         -25.1$ & $        0.201$ & $    \bm{-24.9}$ & $\bigcirc$ \\
            QKP\_GEN\_004             & 1 / 0 & $      \sim 0.0$ & $         -9.98$ & $        0.211$ & $    \bm{-9.77}$ & $\bigcirc$ \\
            QKP\_GEN\_005             & 0 / 0 & $           0.0$ & $         -15.3$ & $        0.188$ & $    \bm{-15.1}$ & $\bigcirc$ \\
            QPLIB\_0633               & 3 / 3 & $           0.0$ & $         0.007$ & $        0.144$ & $         0.151$ &            \\
            QPLIB\_3750               & 3 / 3 & $           0.0$ & $        -0.036$ & $        0.141$ & $         0.105$ &            \\
            QPLIB\_3832               & 2 / 2 & $           0.0$ & $        -0.029$ & $        0.144$ & $         0.114$ &            \\
            QPLIB\_6647               & 2 / 2 & $           0.0$ & $         0.101$ & $        0.141$ & $         0.242$ &            \\
            QPLIB\_10048              & 2 / 3 & $           0.0$ & $         0.016$ & $        0.146$ & $         0.162$ &            \\
            QPLIB\_10058              & 3 / 3 & $           0.0$ & $        -0.168$ & $        0.140$ & $        -0.028$ &            \\
            QPLIB\_10061              & 2 / 3 & $           0.0$ & $        -0.103$ & $        0.111$ & $         0.008$ &            \\
            QPLIB\_10065              & 3 / 3 & $           0.0$ & $         0.004$ & $        0.135$ & $         0.139$ &            \\
            QPLIB\_10067              & 3 / 3 & $           0.0$ & $        -0.023$ & $        0.114$ & $         0.090$ &            \\
            TSP\_GEN\_000             & 1 / 1 & $    \bm{-1.07}$ & $         -2.80$ & $        0.137$ & $    \bm{-2.67}$ & $\bigcirc$ \\
            TSP\_GEN\_001             & 1 / 1 & $    \bm{-2.66}$ & $         -8.98$ & $        0.151$ & $    \bm{-8.83}$ & $\bigcirc$ \\
            TSP\_GEN\_002             & 1 / 1 & $    \bm{-5.17}$ & $         -1.73$ & $        0.147$ & $    \bm{-1.59}$ & $\bigcirc$ \\
            TSPLIB\_burma14           & 2 / 2 & $           0.0$ & $        -0.060$ & $        0.121$ & $         0.061$ &            \\
            TSPLIB\_hk48              & 1 / 1 & $    \bm{-1.76}$ & $         -7.97$ & $        0.151$ & $    \bm{-7.82}$ & $\bigcirc$ \\
            MIPLIB\_cvs16r128-89      & - / 2 & $    \bm{-5.78}$ & $         -3.66$ & $        0.041$ & $    \bm{-3.62}$ & $\bigcirc$ \\
            \parbox{0.2\linewidth}{MIPLIB\_neos-2328163-agri} & - / 2 & $        -0.184$ & $         -6.59$ & $        0.041$ & $    \bm{-6.55}$ & $\bigcirc$ \\
            MIPLIB\_seymour           & - / 2 & $        -0.539$ & $         -3.14$ & $        0.041$ & $    \bm{-3.10}$ & $\bigcirc$ \\
            MIPLIB\_queens-30         & - / 2 & $             0$ & $         -1.01$ & $        0.041$ & $        -0.969$ &            \\
            MIPLIB\_reblock354        & - / 3 & $             0$ & $         -17.9$ & $        0.051$ & $    \bm{-17.8}$ & $\bigcirc$ \\
            \hline
        \end{tabular}
        \caption{
            Comparison of the results from running DA on test data with classified and selected parameters and DA with default parameters.
            The "Instance ID" column is an instance ID.
            In the column "GNN(\C{X})," True is the true value, and Pred is the predicted value in the classification result by GNN.
            Since the true value of MIPLIB is introduced as out-of-distribution data, the true value is not indicated, and a horizontal line (-) is drawn.
            The column "GAP" is the value of $100 \times \frac{E^{(class)} - E^{(default)}}{|E^{(default)}|}$, and the column "$\bm{\Delta} TTS$" is the value of $TTS^{(class)} - TTS^{(default)}$.
            The "$T_{tune}$" column represents the processing time taken when predicting by GNN, and the "$\bm{\Delta} T_{tot}$" column is the value of $\bm{\Delta} TTS + T_{tune}$.
            The "Effect" column shows user's perception of improvement or deterioration as $\bigcirc\ /\ \times$.
            }
        \label{tb:cost_gap_last_update_time_diff_default}
    \end{table}
}

{
    \tabcolsep=2pt

    \begin{table}[htbp]
        \centering\scriptsize
        \begin{tabular}{|l|c|r|r|r|r|c|} \hline
            \multirow{2}{*}{Instance ID} & GNN(\C{X}) & \multicolumn{1}{|c}{Gap} & \multicolumn{1}{|c}{$\Delta TTS$} & \multicolumn{1}{|c}{$T_{tune}$} & \multicolumn{1}{|c|}{$\Delta T_{tot}$} & \multirow{2}{*}{Effect} \\
            & True/Pred & \multicolumn{1}{|c}{[\%]} & \multicolumn{1}{|c}{[s]} & \multicolumn{1}{|c}{[s]} & \multicolumn{1}{|c|}{[s]} & \\ \hline
            QKP\_GEN\_000             & 0 / 0 & $     0.001$ & $   7.43$ & $ 0.211$ & $      7.64$ &   $\times$ \\
            QKP\_GEN\_001             & 0 / 0 & $       0.0$ & $ -0.457$ & $ 0.145$ & $    -0.312$ &            \\
            QKP\_GEN\_002             & 0 / 0 & $       0.0$ & $  0.489$ & $ 0.169$ & $     0.658$ &            \\
            QKP\_GEN\_003             & 0 / 0 & $       0.0$ & $  0.467$ & $ 0.201$ & $     0.669$ &            \\
            QKP\_GEN\_004             & 1 / 0 & $  \sim 0.0$ & $  -2.42$ & $ 0.211$ & $\bm{-2.21}$ & $\bigcirc$ \\
            QKP\_GEN\_005             & 0 / 0 & $       0.0$ & $ -0.598$ & $ 0.188$ & $    -0.409$ &            \\
            QPLIB\_0633               & 3 / 3 & $       0.0$ & $  0.019$ & $ 0.144$ & $     0.163$ &            \\
            QPLIB\_3750               & 3 / 3 & $       0.0$ & $ -0.144$ & $ 0.141$ & $    -0.002$ &            \\
            QPLIB\_3832               & 2 / 2 & $    -0.911$ & $  -6.61$ & $ 0.144$ & $\bm{-6.46}$ & $\bigcirc$ \\
            QPLIB\_6647               & 2 / 2 & $       0.0$ & $  -1.56$ & $ 0.141$ & $\bm{-1.42}$ & $\bigcirc$ \\
            QPLIB\_10048              & 2 / 3 & $       0.0$ & $  0.069$ & $ 0.146$ & $     0.215$ &            \\
            QPLIB\_10058              & 3 / 3 & $       0.0$ & $ -0.062$ & $ 0.140$ & $     0.078$ &            \\
            QPLIB\_10061              & 2 / 3 & $       0.0$ & $ -0.094$ & $ 0.111$ & $     0.018$ &            \\
            QPLIB\_10065              & 3 / 3 & $       0.0$ & $ -0.088$ & $ 0.135$ & $     0.048$ &            \\
            QPLIB\_10067              & 3 / 3 & $       0.0$ & $ -0.074$ & $ 0.114$ & $     0.040$ &            \\
            TSP\_GEN\_000             & 1 / 1 & $     0.013$ & $   8.06$ & $ 0.137$ & $      8.19$ &   $\times$ \\
            TSP\_GEN\_001             & 1 / 1 & $     0.376$ & $  0.323$ & $ 0.151$ & $     0.475$ &            \\
            TSP\_GEN\_002             & 1 / 1 & $     0.708$ & $ -0.074$ & $ 0.147$ & $     0.073$ &            \\
            TSPLIB\_burma14           & 2 / 2 & $       0.0$ & $  0.051$ & $ 0.121$ & $     0.171$ &            \\
            TSPLIB\_hk48              & 1 / 1 & $     0.547$ & $  -2.49$ & $ 0.151$ & $\bm{-2.34}$ & $\bigcirc$ \\
            MIPLIB\_cvs16r128-89      & - / 2 & $\bm{-15.7}$ & $  0.228$ & $ 0.041$ & $     0.269$ &            \\
            \parbox{0.2\linewidth}{MIPLIB\_neos-2328163-agri} & - / 2 &            - &         - & $ 0.041$ &            - & $\bigcirc$ \\
            MIPLIB\_seymour           & - / 2 & $\bm{-2.08}$ & $  -1.12$ & $ 0.041$ & $\bm{-1.08}$ & $\bigcirc$ \\
            MIPLIB\_queens-30         & - / 2 & $\bm{-1.04}$ & $  -1.69$ & $ 0.041$ & $\bm{-1.65}$ & $\bigcirc$ \\
            MIPLIB\_reblock354        & - / 3 & $       0.0$ & $  -10.2$ & $ 0.051$ & $\bm{-10.2}$ & $\bigcirc$ \\
            \hline
        \end{tabular}
        \caption{
            Comparison of classification results with those run with parameters tuned for each problem.
            The columns "instance ID," "GNN(\C{X})," and "Gap" are the same as those in Table.~\ref{tb:cost_gap_last_update_time_diff_default}.
            Columns "Gap," "$\bm{\Delta} TTS$," and "$\bm{\Delta} T_{tot}$" are the corresponding columns in Table.~\ref{tb:cost_gap_last_update_time_diff_default} that replace the results of the default parameters with the those of the parameters tuned for each problem.
            The values of $\mathrm{Gap}$, $\bm{\Delta} TTS$, and $\bm{\Delta} T_{tot}$ in MIPLIB1\_neos-2328163-agri are indicated by a horizontal line (-) because no satisfactory solution could be obtained from executing with parameters tuned for each problem.
            }
        \label{tb:cost_gap_last_update_time_diff_problem}
    \end{table}
}

\section{Conclusion} \label{sec:conclusion}

In practical applications using a CO solver, the execution time is represented by the sum of the $TSS$ and the $T_{tune}$, $TSS+T_{tune}$.
Minimizing this total is beneficial for users.
The ISAC method was developed to learn and select suitable solver parameters for various CO problem instances, eliminating the need to tune individual instances.
However, even with ISAC, if the features including those with high computational costs, $T_{tune}$ can become significant and cannot be ignored.
In this study, we conduct the computationally intensive feature extraction only during the training step and replace it with the computationally lightweight GNN classification in the execution step.
We conducted a demonstration using the digital annealer, a heuristic solver specialized for BQP problems developed by Fujitsu.
The computationally intensive features used were summary statistics from the update history and feasible solutions extracted from the solver log during a $30\ \mathrm{s}$ run.
As a result of training the GNN based on the training data set consisting of TSP, QKP, and QPLIB, we succeeded in reducing the solver execution time for feature extraction from $10\ \mathrm{seeds} \times 30\ \mathrm{s} =300\ \mathrm{s}$ of the ISAC execution step in Fig.~\ref{fig:Fig_ISAC}-(B) to sub-seconds for GNN prediction in Fig.~\ref{fig:Fig_fastISACwGNN}-(B), achieving a classification accuracy of over $85\ \%$.
When evaluating 25 instances, including MIPLIB that is an out-of-distribution data set relative to the training data set, the method proved effective in nearly all TSP, QKP, and MIPLIB cases compared to default parameters.
Although the evaluation runtime was $30\ \mathrm{s}$, certain instances like QKP showed significant improvements, with a reduction in time-to-reach the solution by more than $20\ \mathrm{s}$, while maintaining nearly identical cost values.
In addition, the evaluation revealed that the method achieves nearly the same level of performance as parameters tuned for each specific problem.
This indicates that the system can adequately predict instance distribution, which are initially unknown, and effectively select the appropriate parameters.

In this experiment, we deliberately omitted simple statistical features such as instance size and density to test whether the GNN could accurately predict classes formed by computationally intensive features directly related to solver performance.
For future work, we aim to develop an ISAC framework that combines simple statistical features and computationally intensive ones.
During the training step, we will cluster features that merge these two types.
In the execution step, we plan to explore various frameworks, such as using GNNs to predict classes, employing a multimodal model combining a GNN for instance graphs and a neural network for statistical feature vectors, and regressing computationally intensive features with GNNs while incorporating simple statistical features to determine nearby clusters.


\bibliographystyle{amsplain}
\bibliography{2024_arXiv_fastISACwGNN}

\end{document}